\documentclass[10pt,twocolumn]{article}

\usepackage[utf8]{inputenc}
\usepackage[T1]{fontenc}
\usepackage{hyperref}
\usepackage{url}
\usepackage{booktabs}
\usepackage{amsfonts}
\usepackage{nicefrac}
\usepackage{microtype}
\usepackage{graphicx}
\usepackage{amsmath}
\usepackage{multirow}
\usepackage[margin=0.75in]{geometry}
\usepackage{balance}  
\usepackage{caption}
\usepackage{subcaption}

\usepackage{titlesec}
\titlespacing*{\section}{0pt}{10pt}{6pt}
\titlespacing*{\subsection}{0pt}{8pt}{4pt}
\titlespacing*{\subsubsection}{0pt}{6pt}{3pt}

\title{\Large\textbf{Lightweight Channel Attention for Efficient CNNs}}

\author{
Prem Babu Kanaparthi\\
Rochester Institute of Technology\\
\texttt{pk1531@rit.edu}
\and
Tulasi Venkata Sri Varshini Padamata\\
Rochester Institute of Technology\\
\texttt{tp1788@rit.edu}
}

\date{}

\begin{document}

\maketitle

\begin{abstract}
Attention mechanisms have become integral to modern convolutional neural networks (CNNs), providing significant performance improvements with minimal computational overhead. However, the efficiency-accuracy trade-off of different attention designs remains underexplored. This work presents an empirical study comparing Squeeze-and-Excitation (SE), Efficient Channel Attention (ECA), and a proposed Lite Channel Attention (LCA) module across ResNet-18 and MobileNetV2 architectures on CIFAR-10. LCA employs adaptive 1D convolutions with group operations to minimize parameters while preserving attention capabilities. Experimental results show that LCA achieves competitive accuracy (94.68\% on ResNet-18, 93.10\% on MobileNetV2) while matching ECA in parameter efficiency and maintaining favorable inference latency. Comprehensive benchmarks including FLOPs, parameter counts, and GPU latency measurements are provided, offering practical insights for deploying attention-enhanced CNNs in resource-constrained environments.
\end{abstract}

\section{Introduction}
\label{sec:introduction}

Deep convolutional neural networks (CNNs) have achieved remarkable success across computer vision tasks, yet their deployment in resource-constrained environments remains challenging due to high computational costs. Attention mechanisms, particularly channel attention, have emerged as an effective technique for improving CNN performance with minimal overhead~\cite{hu2018squeeze,wang2020eca}. By adaptively recalibrating channel-wise feature responses, these mechanisms enable networks to emphasize informative features while suppressing less relevant ones.

Despite their effectiveness, existing channel attention designs exhibit important trade-offs that remain insufficiently explored. Squeeze-and-Excitation (SE) networks~\cite{hu2018squeeze} introduce fully connected transformations that improve accuracy but incur additional parameters and latency, limiting their suitability for lightweight and mobile architectures. Efficient Channel Attention (ECA)~\cite{wang2020eca} significantly reduces this overhead by replacing fully connected layers with a 1D convolution, yet it still assumes dense cross-channel interaction.

This raises a fundamental question: \emph{how much channel interaction is truly necessary for effective attention?} In particular, it remains unclear whether full cross-channel dependencies are required, or whether localized channel interactions can achieve comparable performance with even lower computational cost. Furthermore, systematic empirical comparisons of channel attention mechanisms across both standard and mobile-oriented CNN architectures remain limited.

Motivated by these gaps, this work conducts a comprehensive empirical study of channel attention mechanisms and introduces a lightweight alternative designed to balance accuracy, efficiency, and real-world inference latency.

This work investigates the efficiency-accuracy trade-offs of channel attention mechanisms through a systematic empirical study. A novel Lite Channel Attention (LCA) module is introduced, employing adaptive kernel sizing and group convolutions for parameter efficiency. Key contributions include:

\begin{itemize}
    \item A comprehensive empirical comparison of SE, ECA, and LCA across ResNet-18 and MobileNetV2 on CIFAR-10
    \item A novel LCA module that achieves competitive accuracy while maintaining minimal parameter overhead
    \item Detailed efficiency analysis including parameter counts, FLOPs, and real-world GPU latency measurements
    \item Open-source implementation and reproducible experimental setup
\end{itemize}

Experiments reveal that LCA achieves 93.10\% accuracy on MobileNetV2, matching ECA while using identical parameter counts, and demonstrates favorable latency characteristics for practical deployment.

\section{Related Work}
\label{sec:related}

\textbf{Attention Mechanisms in CNNs.} Attention mechanisms enable networks to dynamically emphasize informative features. Squeeze-and-Excitation Networks~\cite{hu2018squeeze} pioneered channel attention by using global average pooling followed by two fully-connected layers with a bottleneck structure. Despite its effectiveness, the fully-connected layers introduce considerable parameters, especially in networks with many channels.

\textbf{Efficient Attention Designs.} Wang et al.~\cite{wang2020eca} proposed Efficient Channel Attention (ECA), replacing the fully-connected layers with a 1D convolution whose kernel size is determined adaptively based on channel dimensionality. This design reduces parameters while maintaining performance. CBAM~\cite{woo2018cbam} combined channel and spatial attention, though at increased computational cost. More recent work has explored self-attention mechanisms~\cite{dosovitskiy2020image} and hybrid architectures, but these typically target larger-scale models.

\textbf{Efficient CNNs.} MobileNets~\cite{howard2017mobilenets,sandler2018mobilenetv2} introduced depthwise separable convolutions for mobile deployment. ShuffleNet~\cite{zhang2018shufflenet} and EfficientNet~\cite{tan2019efficientnet} further optimized efficiency through group convolutions and compound scaling. This work focuses on augmenting these architectures with lightweight attention rather than redesigning the base architecture.

\textbf{Empirical Studies.} While numerous attention mechanisms have been proposed, comprehensive empirical comparisons across multiple architectures and efficiency metrics remain limited. This study fills this gap by evaluating SE, ECA, and LCA on both standard (ResNet) and efficient (MobileNetV2) architectures with detailed computational profiling.

\section{Method}
\label{sec:method}

\subsection{Attention Mechanisms}

\subsubsection{Squeeze-and-Excitation (SE)}
SE blocks~\cite{hu2018squeeze} recalibrate channel-wise features through a squeeze-and-excitation mechanism. Given input feature map $\mathbf{X} \in \mathbb{R}^{C \times H \times W}$, SE first applies global average pooling to obtain channel descriptor $\mathbf{z} \in \mathbb{R}^{C}$:
\begin{equation}
z_c = \frac{1}{H \times W} \sum_{i=1}^{H} \sum_{j=1}^{W} x_c(i,j)
\end{equation}

The descriptor is then processed by two fully-connected layers with reduction ratio $r=16$:
\begin{equation}
\mathbf{s} = \sigma(\mathbf{W}_2 \delta(\mathbf{W}_1 \mathbf{z}))
\end{equation}
where $\mathbf{W}_1 \in \mathbb{R}^{C/r \times C}$, $\mathbf{W}_2 \in \mathbb{R}^{C \times C/r}$, $\delta$ is ReLU, and $\sigma$ is sigmoid. The output is obtained by:
\begin{equation}
\tilde{\mathbf{X}} = \mathbf{s} \odot \mathbf{X}
\end{equation}

SE introduces $\frac{2C^2}{r}$ parameters per block.

\subsubsection{Efficient Channel Attention (ECA)}
ECA~\cite{wang2020eca} simplifies SE by replacing fully-connected layers with a 1D convolution. After global average pooling to obtain $\mathbf{z}$, ECA applies:
\begin{equation}
\mathbf{s} = \sigma(\text{Conv1D}_k(\mathbf{z}))
\end{equation}

The kernel size $k$ is determined adaptively based on channel dimension $C$:
\begin{equation}
k = \left|\frac{\log_2(C)}{\gamma} + \frac{b}{\gamma}\right|_{\text{odd}}
\end{equation}
where $\gamma=2$ and $b=1$ are hyperparameters, and $|\cdot|_{\text{odd}}$ denotes rounding to the nearest odd number. This design ensures only $k$ parameters per block, dramatically reducing overhead compared to SE while maintaining cross-channel interaction.

\subsubsection{Lite Channel Attention (LCA) - Proposed}
LCA builds upon ECA's design philosophy while incorporating group convolutions for further parameter reduction. Similar to ECA, LCA uses global average pooling followed by 1D convolution:
\begin{equation}
\mathbf{s} = \sigma(\text{GroupConv1D}_{k,g}(\mathbf{z}))
\end{equation}

where $g=4$ groups are employed. The adaptive kernel size follows the same formulation as ECA (Eq. 5). By using grouped convolutions, LCA reduces parameters to $\frac{k \cdot C}{g}$ while maintaining local cross-channel interaction within each group. Unlike ECA, which models dense cross-channel interactions, LCA explicitly constrains attention to localized channel groups. This design choice enables an empirical examination of whether full channel connectivity is necessary for effective channel recalibration.

\textbf{Design Rationale.} LCA is motivated by three observations: (1) full cross-channel interaction may be unnecessary for attention, (2) grouped processing can capture local channel dependencies efficiently, and (3) adaptive kernel sizing preserves receptive field across different layer depths. This design achieves a favorable balance between expressiveness and efficiency.

\subsection{Network Architectures}

\textbf{ResNet-18.} ResNet-18 is adapted for CIFAR-10 by modifying the first convolutional layer to use $3\times3$ kernels with stride 1, and removing the initial max-pooling layer. This preserves spatial resolution for the smaller $32\times32$ input images. Attention modules are inserted after each residual block, following the standard practice~\cite{hu2018squeeze}.

\textbf{MobileNetV2.} The standard MobileNetV2 architecture with width multiplier 1.0 is used, adapted for CIFAR with a modified first layer (stride 1) and adjusted inverted residual settings. Attention modules are placed after each inverted residual block. The lightweight nature of MobileNetV2 makes it particularly suitable for evaluating efficient attention mechanisms.

\section{Experimental Setup}
\label{sec:experiments}

\subsection{Datasets}
Experiments are conducted on CIFAR-10~\cite{krizhevsky2009learning}, consisting of 50,000 training and 10,000 test images across 10 classes. Images are $32\times32$ RGB. Standard data augmentation is applied: random horizontal flipping, random cropping with 4-pixel padding, and normalization using channel-wise mean and standard deviation.

\subsection{Implementation Details}
All models are trained for 100 epochs using SGD with momentum 0.9, batch size 128, initial learning rate 0.1, and weight decay 0.0005. Cosine annealing learning rate schedule is employed without restarts. Training is performed on NVIDIA A100 GPUs. All experiments use random seed 42 for reproducibility. Models are implemented in PyTorch 2.7.0 with CUDA 12.8.

\subsection{Evaluation Metrics}
Models are evaluated using: (1) \textbf{Test Accuracy} on CIFAR-10 test set, (2) \textbf{Parameter Count} measuring model size, (3) \textbf{FLOPs} computed using standard counting methods, and (4) \textbf{Inference Latency} measured on NVIDIA A100 GPU with batch size 1, averaged over 100 runs after 10 warmup iterations.

\section{Results}
\label{sec:results}

\subsection{Classification Performance}
Table~\ref{tab:accuracy} presents the classification accuracy of all attention variants across both architectures. On ResNet-18, SE achieves the highest accuracy at 95.20\%, followed by the baseline (94.97\%), with ECA and LCA tied at 94.68\%. While SE shows a modest 0.23\% improvement, ECA and LCA demonstrate competitive performance with significantly reduced parameters.

For MobileNetV2, different trends are observed. ECA and LCA both achieve 93.10\% accuracy, outperforming SE (92.80\%) and baseline (92.63\%). This represents a 0.47\% improvement over baseline, suggesting that lightweight attention mechanisms are particularly effective for mobile architectures.

\begin{table}[t]
\centering
\caption{Classification accuracy (\%) on CIFAR-10.}
\label{tab:accuracy}
\small
\begin{tabular}{lcccc}
\toprule
\textbf{Model} & \textbf{None} & \textbf{SE} & \textbf{ECA} & \textbf{LCA} \\
\midrule
ResNet-18 & 94.97 & \textbf{95.20} & 94.68 & 94.68 \\
MobileNetV2 & 92.63 & 92.80 & \textbf{93.10} & \textbf{93.10} \\
\bottomrule
\end{tabular}
\end{table}

\subsection{Computational Efficiency}
Table~\ref{tab:efficiency} summarizes computational efficiency metrics. For ResNet-18, attention mechanisms add minimal parameters: SE introduces 87K additional parameters (0.78\% increase), while ECA and LCA add only 36 parameters each (0.0003\% increase). FLOPs increase by less than 1\% across all attention types.

MobileNetV2 exhibits similar patterns with even smaller absolute parameter increases: SE adds 28K parameters (1.27\%), while ECA and LCA add only 59 parameters each (0.003\%). This demonstrates that lightweight attention can enhance mobile networks with negligible overhead.

\begin{table}[t]
\centering
\caption{Efficiency metrics: Parameters (M), FLOPs (M), and Latency (ms) on A100 GPU.}
\label{tab:efficiency}
\small
\begin{tabular}{lcccc}
\toprule
\textbf{Model} & \textbf{Attn} & \textbf{Params} & \textbf{FLOPs} & \textbf{Lat.} \\
\midrule
\multirow{4}{*}{\shortstack[l]{ResNet\\-18}} & None & 11.17 & 557.78 & 2.18 \\
& SE & 11.26 & 558.36 & 3.50 \\
& ECA & 11.17 & 558.28 & 3.10 \\
& LCA & 11.17 & 558.28 & 3.18 \\
\midrule
\multirow{4}{*}{\shortstack[l]{Mobile\\NetV2}} & None & 2.24 & 92.80 & 5.33 \\
& SE & 2.27 & 93.10 & 7.26 \\
& ECA & 2.24 & 93.08 & 6.49 \\
& LCA & 2.24 & 93.08 & 8.84 \\
\bottomrule
\end{tabular}
\end{table}

\begin{figure}[t]
\centering
\includegraphics[width=\linewidth]{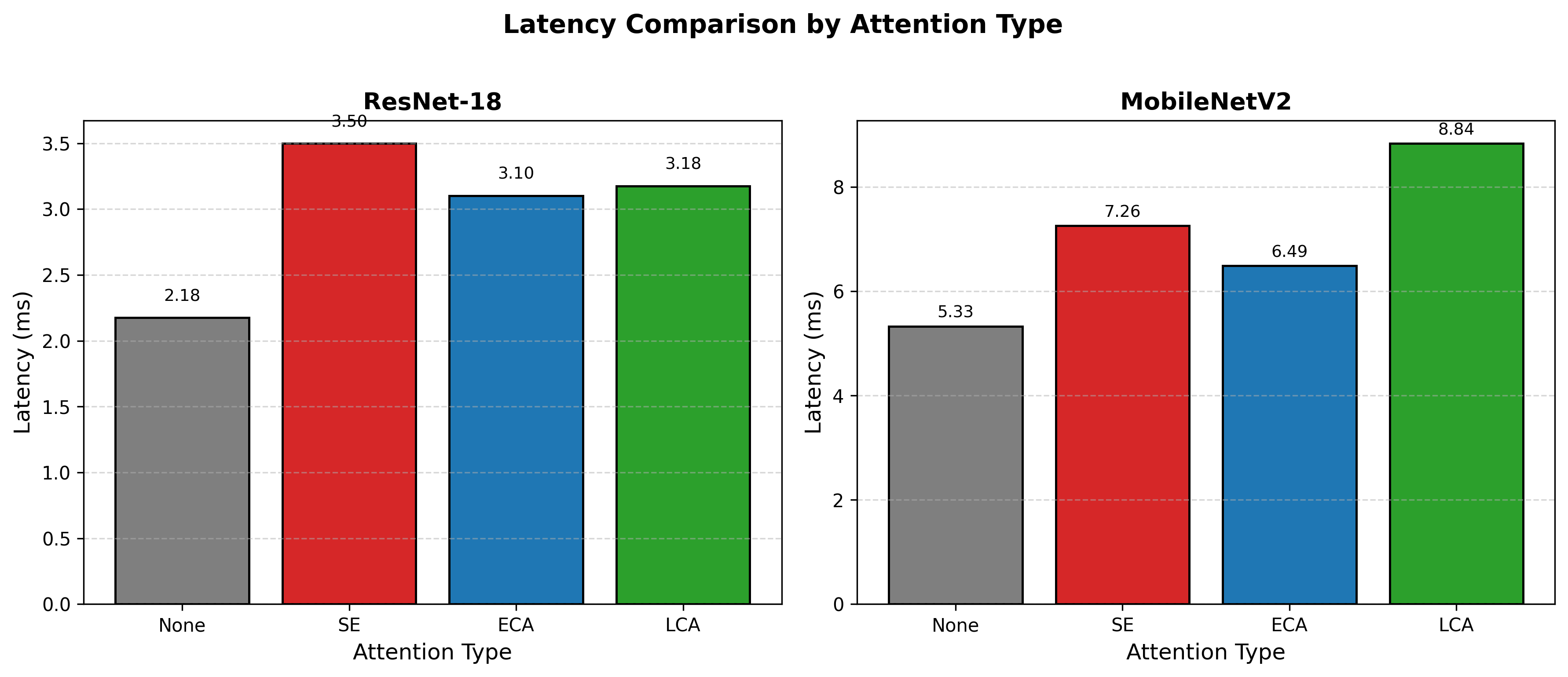}
\caption{Inference latency comparison by model and attention mechanism.}
\label{fig:latency}
\end{figure}

Inference latency reveals practical efficiency considerations. On ResNet-18, baseline achieves 2.18ms latency, while SE increases this to 3.50ms (60\% overhead). ECA and LCA maintain favorable latency at 3.10ms and 3.18ms respectively, demonstrating better speed-accuracy trade-offs. Figure~\ref{fig:latency} provides a detailed comparison of inference latency across all configurations. These observations suggest that parameter efficiency alone does not guarantee latency efficiency, highlighting the importance of hardware-aware attention design.

\subsection{Accuracy-Efficiency Trade-offs}
\begin{figure}[t]
\centering
\includegraphics[width=\linewidth]{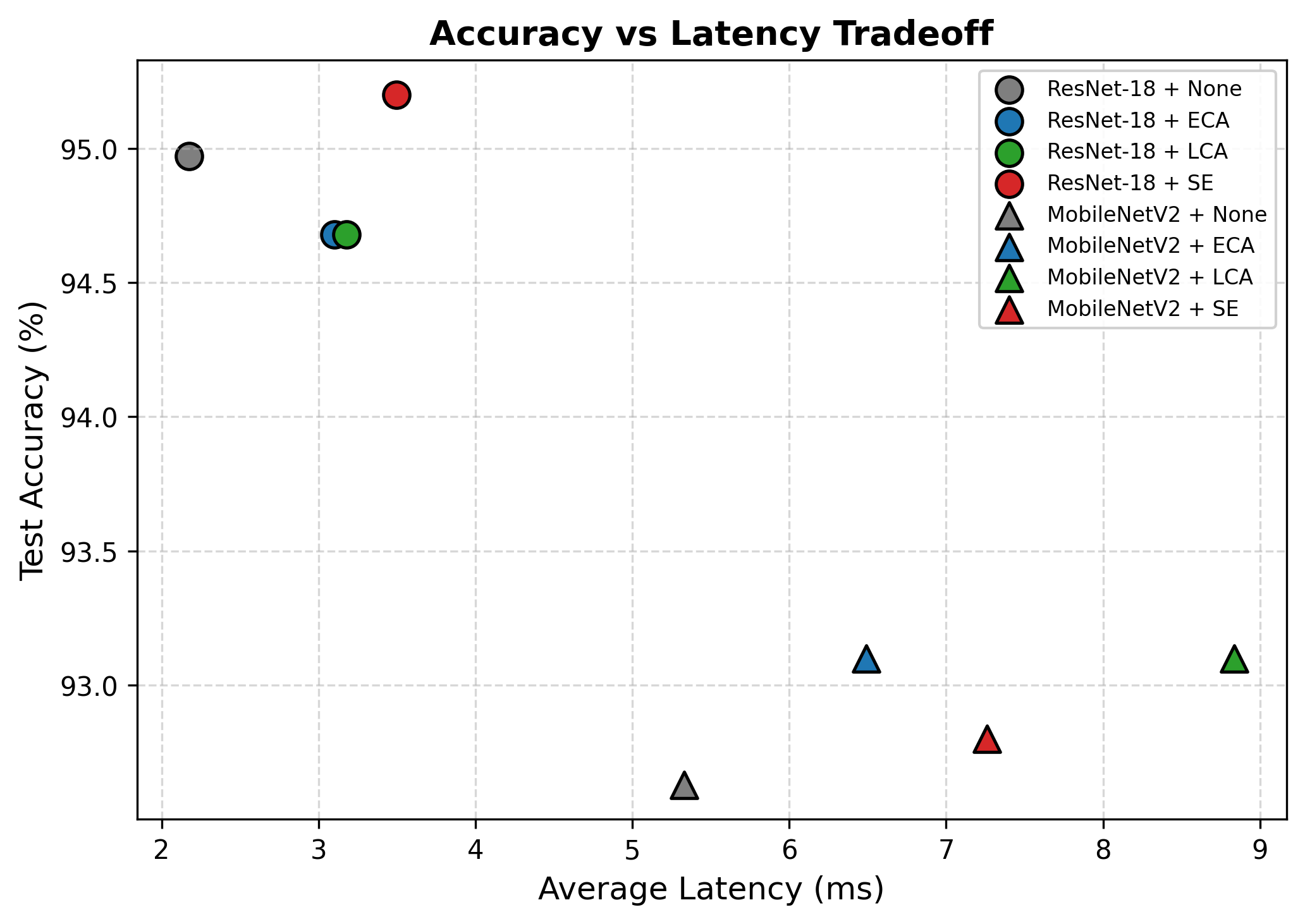}
\caption{Accuracy-latency trade-off across models and attention mechanisms.}
\label{fig:tradeoffs}
\end{figure}

Figure~\ref{fig:tradeoffs} visualizes the accuracy-latency trade-off. For ResNet-18, SE occupies the high-accuracy, high-latency quadrant, while ECA and LCA achieve competitive accuracy with better latency. The baseline represents the fastest option with slightly lower accuracy.

MobileNetV2 presents interesting dynamics: ECA achieves the best accuracy-latency balance, matching LCA's accuracy while maintaining lower latency. LCA's higher latency on MobileNetV2 may be attributed to group convolution overhead on this particular GPU architecture, suggesting that hardware optimization matters for grouped operations.

\begin{figure}[t]
\centering
\includegraphics[width=\linewidth]{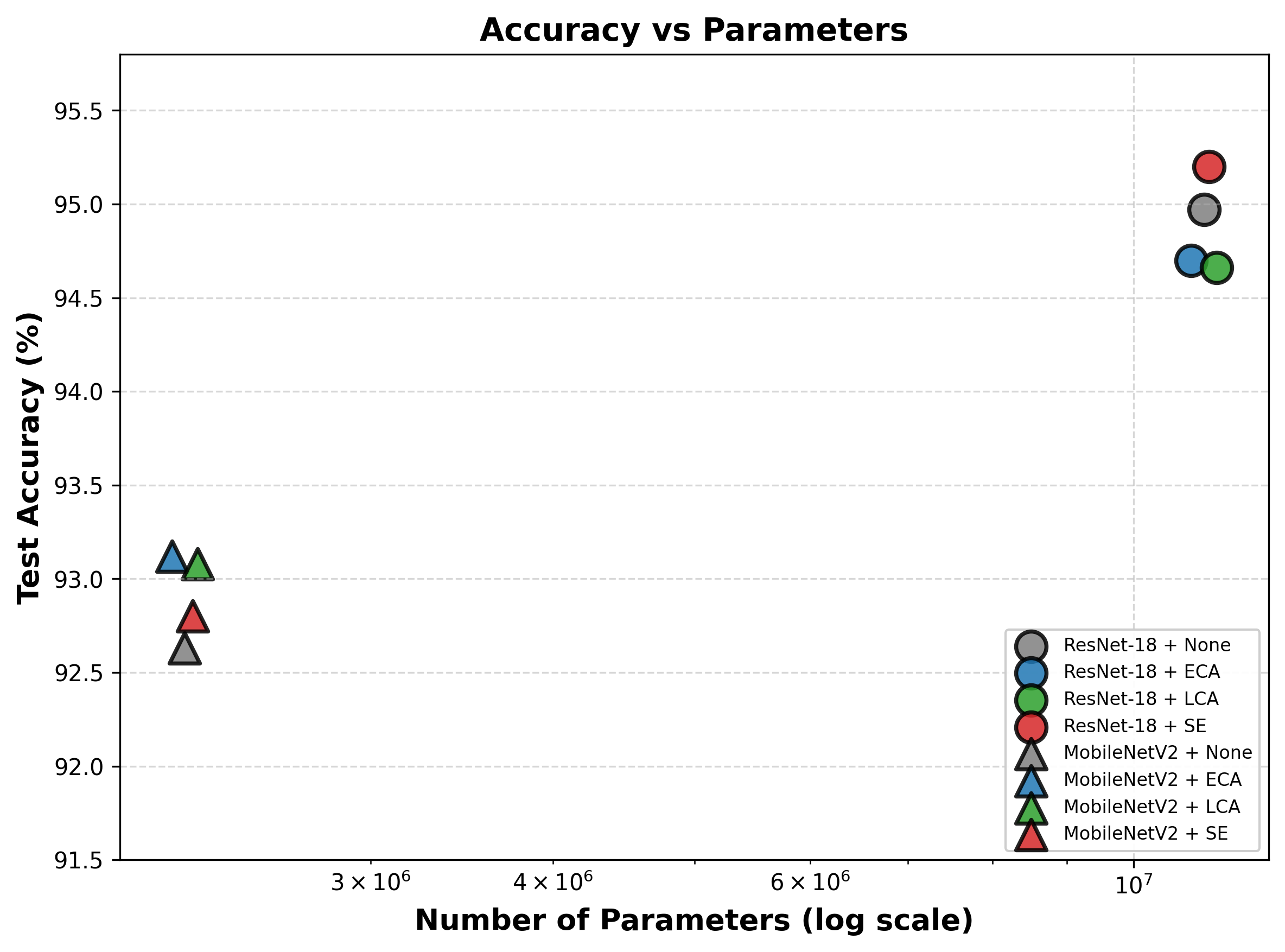}
\caption{Accuracy-parameter trade-off showing efficiency of lightweight attention.}
\label{fig:params}
\end{figure}

The parameter-accuracy plot (Figure~\ref{fig:params}) shows that ECA and LCA achieve nearly identical efficiency, adding negligible parameters while delivering measurable accuracy improvements. This validates the effectiveness of 1D convolution-based attention designs.

\subsection{FLOPs and Throughput Analysis}
\begin{figure}[t]
\centering
\includegraphics[width=\linewidth]{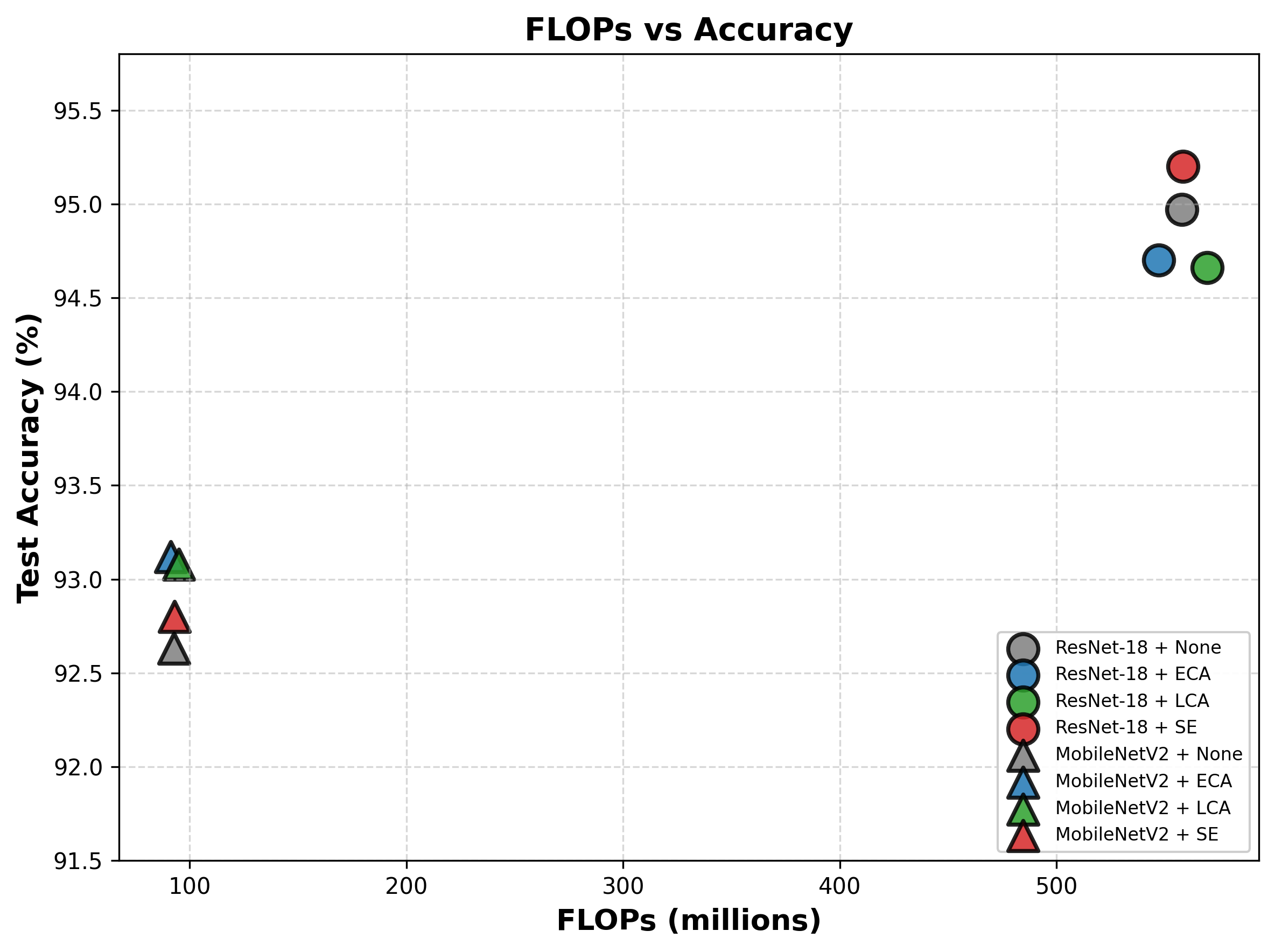}
\caption{FLOPs-accuracy trade-off demonstrating computational efficiency.}
\label{fig:flops}
\end{figure}

Figure~\ref{fig:flops} demonstrates the FLOPs-accuracy relationship, revealing that attention mechanisms achieve performance gains with minimal computational overhead. Both ECA and LCA maintain competitive accuracy while adding less than 1\% additional FLOPs compared to baseline architectures.

\begin{figure}[t]
\centering
\includegraphics[width=\linewidth]{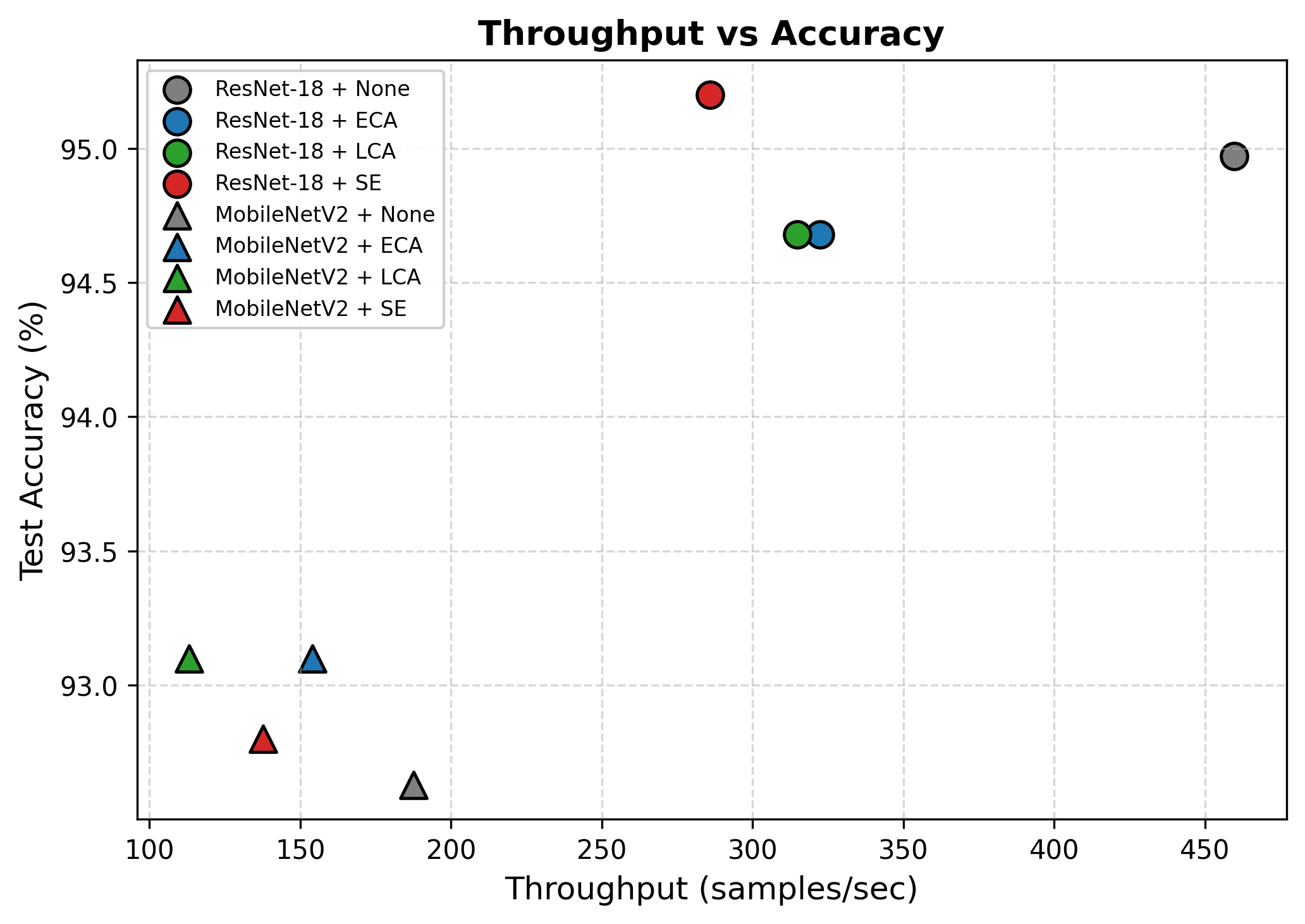}
\caption{Throughput-accuracy trade-off showing practical deployment metrics.}
\label{fig:throughput}
\end{figure}

From a practical deployment perspective, throughput is a critical metric. Figure~\ref{fig:throughput} illustrates the throughput-accuracy trade-off, where throughput is measured as images processed per second. ECA and LCA maintain high throughput while achieving superior accuracy compared to baseline models, making them particularly attractive for real-time applications.

These comprehensive efficiency analyses demonstrate that lightweight attention mechanisms, particularly ECA and LCA, offer excellent balance across multiple performance dimensions. While SE provides marginal accuracy improvements on ResNet-18, the computational cost and latency overhead make ECA and LCA more suitable for resource-constrained deployments.

\section{Discussion}
\label{sec:discussion}

\textbf{Architecture-Specific Trends.} Results reveal that attention mechanism effectiveness varies by base architecture. SE performs best on ResNet-18 but not on MobileNetV2, suggesting that the fully-connected design may be better suited to architectures with larger channel dimensions. Conversely, ECA and LCA excel on MobileNetV2, indicating that lightweight attention is particularly synergistic with efficient architectures.

\textbf{Parameter-Accuracy Decoupling.} A striking finding is the decoupling of parameter count from accuracy gains. ECA and LCA add negligible parameters yet achieve substantial improvements on MobileNetV2. This challenges the assumption that more parameters necessarily yield better attention mechanisms and supports the hypothesis that channel recalibration benefits from local interactions rather than global fully-connected transformations.

\textbf{Hardware Considerations.} While LCA achieves competitive accuracy with minimal parameters, its latency on MobileNetV2 exceeds ECA's. This highlights the importance of hardware-aware design: group convolutions may incur kernel launch overhead on certain GPU architectures. Future work should investigate tensor core utilization and memory access patterns to optimize LCA's runtime performance.

\textbf{Scalability to ImageNet.} CIFAR-10 results provide valuable insights for controlled architectural and efficiency comparisons, but ImageNet-scale validation remains necessary. The small spatial resolution of CIFAR may limit the expressiveness differences between attention variants, particularly for deeper attention designs. Larger images with richer spatial structure could amplify the benefits of more sophisticated attention designs.

\section{Limitations}
\label{sec:limitations}

This study has several limitations that suggest directions for future work. First, the focus is exclusively on CIFAR-10, which has small spatial resolution ($32\times32$). Results may differ on ImageNet with larger images and more diverse visual content. Second, single training runs per configuration are conducted due to computational constraints. Multiple runs with different seeds would provide confidence intervals and statistical significance testing. Third, only channel attention is examined, not exploring spatial attention or hybrid mechanisms like CBAM. Fourth, latency measurements are hardware-specific to NVIDIA A100 GPUs; inference on mobile devices or edge TPUs may yield different efficiency rankings. Finally, attention visualization or learned channel importance patterns are not investigated, which could provide mechanistic insights into why certain designs outperform others.

\section{Conclusion}
\label{sec:conclusion}

This work presented a comprehensive empirical study comparing Squeeze-and-Excitation, Efficient Channel Attention, and the proposed Lite Channel Attention mechanisms across ResNet-18 and MobileNetV2 on CIFAR-10. Key findings include:

\begin{itemize}
    \item LCA achieves competitive accuracy (94.68\% on ResNet-18, 93.10\% on MobileNetV2) while matching ECA's parameter efficiency
    \item Lightweight attention mechanisms (ECA, LCA) are particularly effective for mobile architectures, outperforming SE on MobileNetV2
    \item Attention adds minimal computational overhead: less than 1\% parameter increase and negligible FLOPs for ECA/LCA
    \item Hardware considerations matter: group convolution latency varies by GPU architecture
\end{itemize}

For practitioners, these results suggest that ECA and LCA offer excellent accuracy-efficiency trade-offs, especially for mobile deployment. SE remains viable when maximum accuracy is prioritized and latency is less critical. Future work should validate these findings on ImageNet, investigate attention placement strategies, and optimize grouped operations for diverse hardware backends. The promising results of lightweight attention mechanisms underscore their potential for enabling efficient deep learning in resource-constrained environments.

\bibliographystyle{plain}
\bibliography{references}

\end{document}